\documentclass[a4paper]{article}

\usepackage{lmodern}
\usepackage[T1]{fontenc}
\usepackage[utf8]{inputenc}
\usepackage{amssymb,amsmath}
\usepackage{todonotes}
\usepackage{caption, subcaption}
\usepackage{url}
\usepackage{hyperref}
\usepackage{longtable}
\usepackage{graphicx}

\usepackage{listings}
\usepackage{xcolor}
\definecolor{codegreen}{rgb}{0,0.6,0}
\definecolor{codegray}{rgb}{0.5,0.5,0.5}
\definecolor{codepurple}{rgb}{0.58,0,0.82}
\definecolor{backcolour}{rgb}{0.95,0.95,0.92}
\lstdefinestyle{mystyle}{
    backgroundcolor=\color{backcolour},   
    commentstyle=\color{codegreen},
    keywordstyle=\color{magenta},
    numberstyle=\tiny\color{codegray},
    stringstyle=\color{codepurple},
    basicstyle=\ttfamily\footnotesize,
    breakatwhitespace=false,         
    breaklines=true,                 
    captionpos=b,                    
    keepspaces=true,                 
    numbers=left,                    
    numbersep=5pt,                  
    showspaces=false,                
    showstringspaces=false,
    showtabs=false,                  
    tabsize=2
}
\usepackage{algorithm} 
\usepackage{algpseudocode}

\usepackage{hyperref}
\hypersetup{unicode=true,
            pdfborder={0 0 0},
            breaklinks=true}
\usepackage{longtable,booktabs}
\IfFileExists{parskip.sty}{%
\usepackage{parskip}
}{
\setlength{\parindent}{0pt}
\setlength{\parskip}{6pt plus 2pt minus 1pt}
}
 \title{MACARONS:\\ A Modular and Open-Sourced \\Automation System for Vertical Farming}    
\usepackage{authblk} 
\author[1]{Vijja Wichitwechkarn}
\author[1]{Charles Fox} 

\affil[1]{School of Computer Science, University of Lincoln, Brayford Pool, LN6 7TS, UK}

\begin{document}
\maketitle

\begin{abstract}
The Modular Automated Crop Array Online System (MACARONS) is an extensible, scalable, open hardware system for plant transport in automated horticulture systems such as vertical farms.  It is specified to move trays of plants up to 1060mm $\times$ 630mm and 12.5kg at a rate of 100mm/s along the guide rails and 41.7mm/s up the lifts, such as between stations for monitoring and actuating plants. The cost for the construction of one grow unit of MACARONS is 144.96USD which equates to 128.85USD/m$^2$ of grow area. The designs are released and meets the requirements of CERN-OSH-W, which includes step-by-step graphical build instructions and can be built by a typical technical person in one day at a cost of 1535.50 USD.  Integrated tests are included in the build instructions are used to validate against the specifications, and we report on a successful build.  Through a simple analysis, we demonstrate that MACARONS can operate at a rate sufficient to automate tray loading/unloading, to reduce labour costs in a vertical farm.

\end{abstract}

\begin{longtable}[]{@{}l@{}}
\begin{minipage}[t]{0.97\columnwidth}\raggedright\strut

\subsection*{Metadata Overview}\label{h.akaipbqoqfs8}
Main design files: https://gitlab.com/owopen-source/macarons

Target group: commercial growers, researchers and hobbyists interested in vertical farming and robotics. 

Skills required: general mechanical assembly, drilling, soldering, 3D-printing, laser-cutting, Linux command line.

Replication: A first build has been completed and tested, meeting the the specification. A second build using the documentation is in progress.

\subsection*{Keywords}\label{h.kdz351yp7g7c}

{Vertical farming, automation, agriculture, robotics, open-source hardware.}

\strut\end{minipage}\tabularnewline
\bottomrule
\end{longtable}

\section{(1) Overview}

\begin{figure}[h!]
    \centering
    \includegraphics[width=1.0\linewidth]{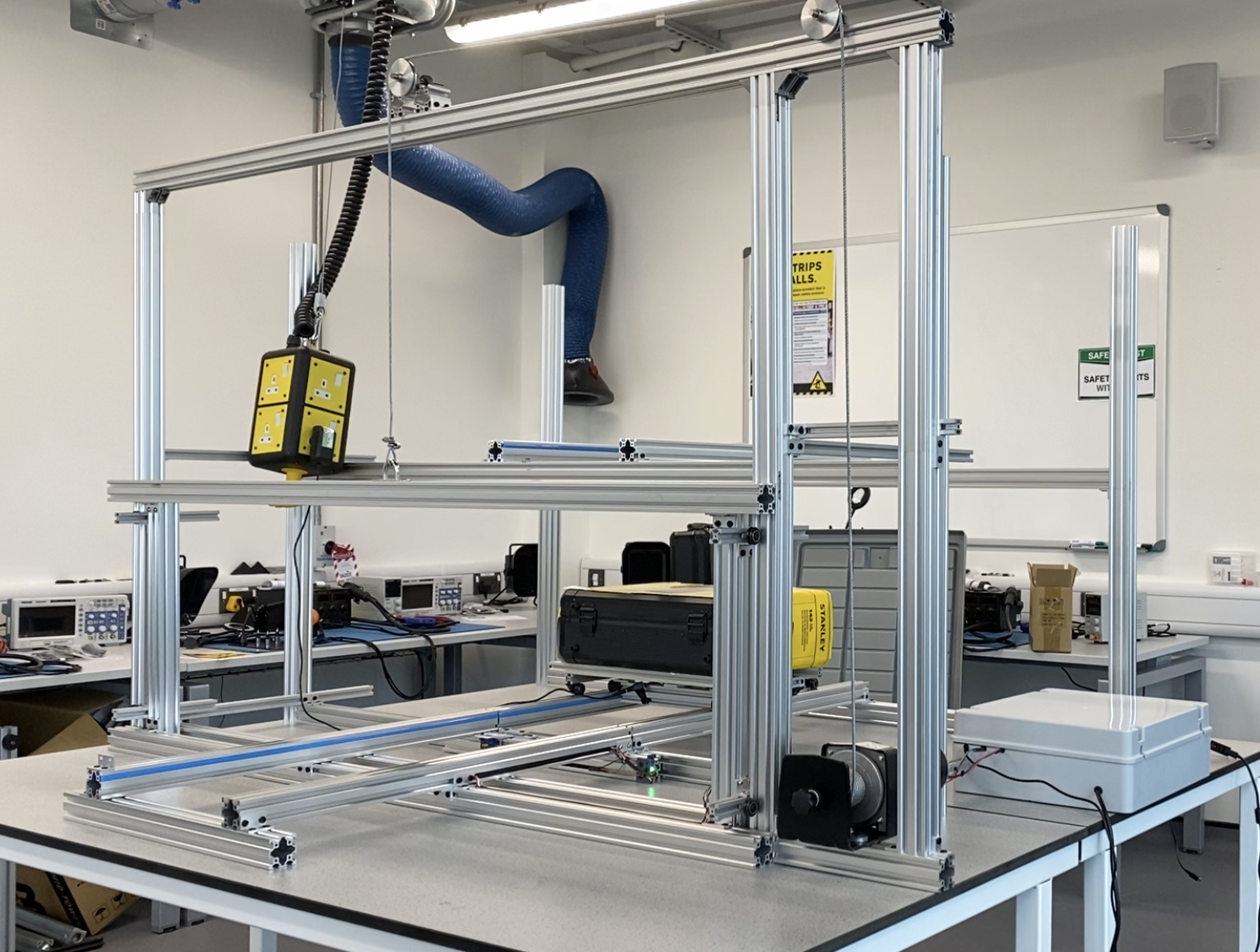}
    \caption{
    Photograph of a 1$\times$2 capacity MACARONS module build.
    }
    \label{fig:macarons_pic}
\end{figure}

Global population is expected to reach 9.7 billion by 2050, with 70\%  expected in urban areas \cite{population}. Without population control interventions, the global amount of food produced must thus  increase by more than 50\%  \cite{50_more_food}. Loss of agricultural land to urban and infrastructure expansion puts further pressure on agriculture  \cite{land_loss}, which is exacerbated by degradation of soil and contamination of water due to uncontrolled use of chemicals, pesticides and herbicides \cite{scarce_land1, scarce_land2}. Climate change is expected to increase the unpredictability of the weather and increase the magnitude and frequency of natural disasters, which will drastically impact the produce from traditional agriculture \cite{future_of_ag}. Agriculture  will have to increase production capacity despite these factors. There is therefore a need for research into more efficient and resilient means of  production \cite{skyward} such as  vertical farms.

Vertical farming involves increasing the effective area utilised to grow crops by stacking crops vertically and inputting electrical energy to provide artificial light in place of or as well as sunlight \cite{future_of_ag}. This is often done in a sealed, indoor, controlled environment setting with hydroponics where the crop is grown directly in a nutrient rich solution without soil. Vertical farming has seen a surge in popularity and attracted recent  attention from both industry and academic research. There is however no standard open-sourced hardware and software platform to support research and development in this area. MACARONS is a first such automated system, shown in Figure \ref{fig:macarons_pic}, that can be scaled and extended to fit a wide variety of space and purposes. The system aims to be a cost-effective and easy to build platform that allows user to automate plant tray transportation. Both software and hardware should be highly customisable, allowing users to adapt it to specific requirements. The system aims to facilitate effective sharing and collaboration to accelerate and accumulate research progress into the optimisation of vertical farms.

Reducing waste throughout the supply chain can also alleviate the increasing pressure on the agriculture sector. A large portion (24\%) of all food never reaches the consumer \cite{24_food} due to long supply chains and inconsistent quality. Transport of fruits and vegetables requiring refrigeration generates twice the amount of carbon dioxide produced by growing them \cite{food_transport_co2}. Vertical farms positioned at key, highly urbanised locations allow for local production of consistent high-quality produce, reducing the amount of transportation required and the amount of produce discarded due to quality. This minimises the cost and carbon emissions associated with transport and refrigeration. Spoilage of the produce through inconsistent cooling during transport is also minimised \cite{transport_cost}. Shorter transport distances also imply longer and more consistent shelf-life, which in turn reduces the amount of produce that goes to waste in the house-hold. Vertical farms may be most useful as part of larger agri-systems which continue to grow in bulk in the countryside but build in options for faster response times and resilience to climate events via their new urban locations.

Although vertical farms avoid direct impact on the local environment such as eutrophication and soil erosion/depletion, they may have much higher energy requirements during growing than traditional agriculture, with the main additional energy requirements being artificial lighting and heating, ventilation and air-conditioning (HVAC) \cite{led_hvac}. Sustainability of vertical farms is therefore debatable, and depends on finding systems in which these costs are lower than those of traditional agricultural and transport processes. Large energy requirements of vertical farms also imply that only small, rapidly grown plants where a large portion of the plant's mass is salable, is currently economically viable in vertical farms \cite{labour_cost}. Important staple crops are currently economically unviable, which is a set back for the promises of alleviating the increase in demand for food from population growth.

Construction of dedicated buildings for vertical farms is expected to result in significant environmental impact \cite{upscale}.  Strategically positioning vertical farms in expensive land in highly urbanised areas come at a high cost, though if not relying on sunlight they can be located underground or in windowless centers of large buildings, which are often underused and available urban space resources. Together with the construction cost, the upfront initial investment for a vertical farm can be over twice that of a high-tech greenhouse that has the same annual production capacity \cite{agritecture}. Operational costs of vertical farms are also very high with a large portion associated with manual labour \cite{labour_cost}. This is exacerbated by high labour costs and scarcity of high skilled labour due to high efficiency requirements.

For vertical farms to live up to their promises, more research and optimisation is needed, especially to reduce their energy consumption. There is therefore a demand for autonomous systems that can facilitate and accelerate this process. Automation will also reduce the operational cost of vertical farms due to labour. An open-sourced standardised platform that can be cost-effectively constructed out of readily available parts would also accelerate the optimisation process as it allows users and researchers to share resources and collaborate more effectively. Open-sourced hardware also provides security for the cost of components and allows for the designs to be forked and modified. To circumvent the construction of dedicated buildings, unused urban spaces such as garages, cellars, basements, ex-industrial sites, offices, retail and residence spaces can be converted into vertical farms. This however requires a system that is flexible enough to be fit to the wide variety of spaces available in the urban fabric, without expensive and time-consuming redesign of custom solutions.

\subsection{Related Systems}
\label{related_systems}

Automation in vertical farms generally includes environmental control and monitoring while transport, loading and unloading of crops into the vertical farm structure is done manually or with the use of large human-operated machines \cite{aerofarms, zcf, atf, bwf}. Using shipping containers to house vertical farms has also been popular. Although fast to deploy in open areas, these solutions are inflexible and cannot be easily incorporated into certain unused urban spaces \cite{gps, ff, ltg}. Some large scale vertical farms automate the transport, loading and unloading of crops with dedicated conveyor belts and robotic systems. This is however expensive to build and highly specific and inflexible. Automated seeding and harvesting is also generally employed at this scale \cite{plenty, tnf}. 

Iron Ox \cite{ironox} is an automated greenhouse where multiple mobile robots move under crop platforms, picking them up and moving them to the watering and harvesting stations. There are also many warehouse robots that use a similar approach  \cite{amazon}. Some systems also use rails to guide their robots such as grocery packing robots \cite{ocado}. The unconstrained environment in the Iron Ox sytem requires advanced localisation and navigation. The system is also not open-source and each robot is expected to be expensive due to the large lifting capacity and sensors. For a comparison a large Amazon warehouse robotic system using a thousand similar robot packers will cost around 15-20 million USD. Each warehousing robot is therefore expected to cost around 20,000USD \cite{amazon_packer_bot}.

The above systems are not open-sourced. FarmBot \cite{farmbot} is an open-sourced project where a fixed size area houses soil for crops to be seeded, watered, weeded and monitored by a robot above it. It is however not built for vertical farming and hydroponics, and is not designed for scaling.  Open Source Ecology \cite{ose} is working on a large ecosystem of open hardware agricultural machines which has inspired MACARONS, but is focused on outdoor farms so does not include vertical farming.

\section{(2) Overall Implementation and Design}\label{h.1u7vph94gfbt}

There is strong demand for an open-sourced, modular, standardised and highly flexible system that can be scaled to fit a wide variety of spaces. The system should be designed with both the constraints from vertical farming and robotics in mind, and automated as far as is reliable. It should also be constructed cheaply out of readily available parts with an online backend that allows for the control, collection of data and analysis in a standardised format.

\begin{figure}[h!]
    \centering
    \includegraphics[width=1.0\linewidth]{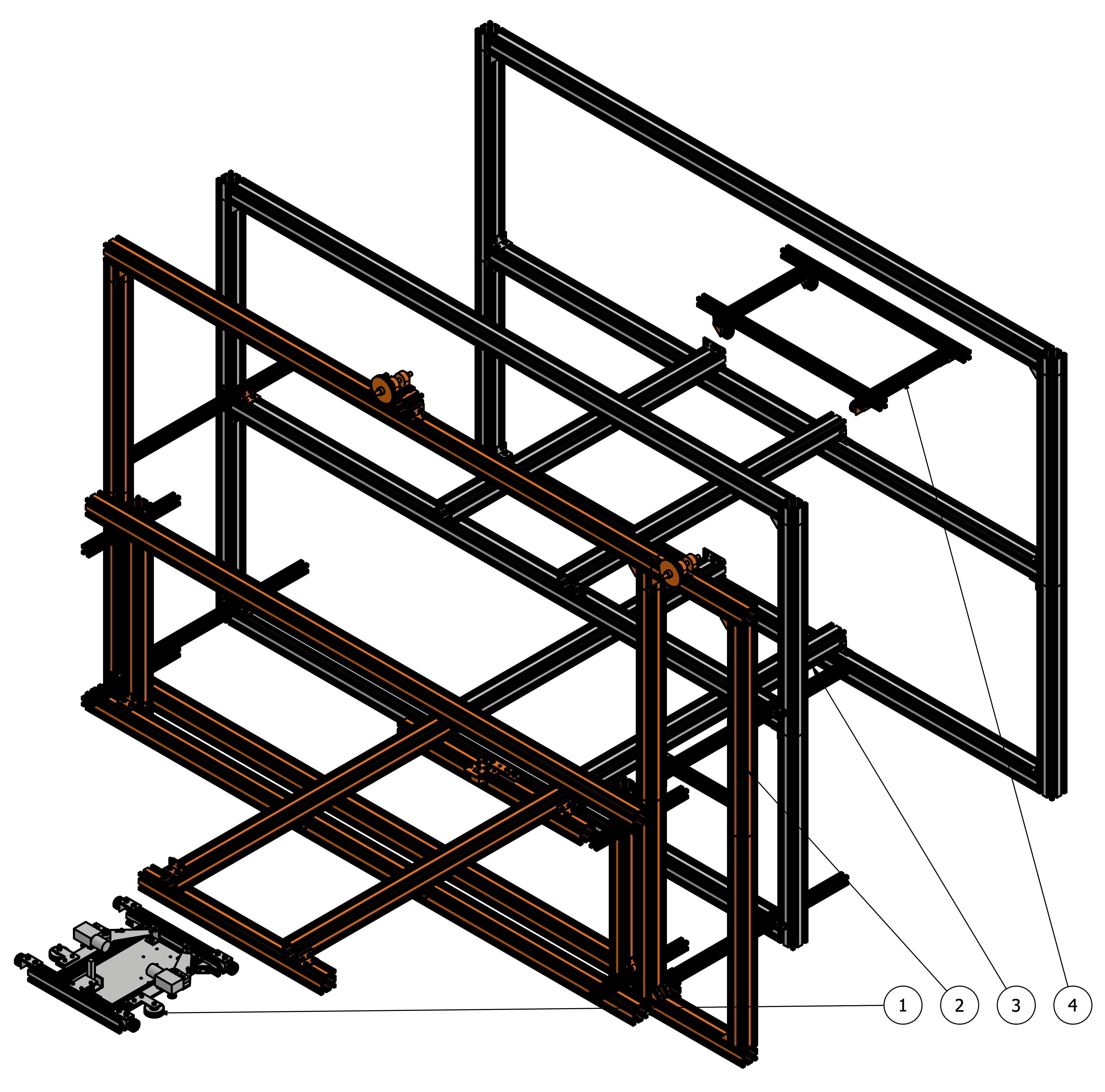}
    \caption{CAD image of a minimal MACARONS. The different components are grouped by colour. Theses include the mover (1), elevator (2), shelves (3) and carriage (4).}
    \label{fig:macarons}
\end{figure}

\subsection{Hardware}
MACARONS (Figure \ref{fig:macarons}) is designed to be modular, scaling up to a variety of sizes as required. Define a \emph{grow unit} as the volume allocated to a single tray of crop that will be grown in the MACARONS. This may eventually be fully occupied by fully grown plants, but begins mostly empty when they are seeds. Define one {\em module} as a self-contained, functioning MACARONS build. The {\em capacity} of a MACARONS module is defined as the number of grow units which can be stacked horizontally ($N_{h}$) and vertically ($N_{v}$) in the system, for instance a $3\times 2$ MACARONS module contains three grow units stacked horizontally and two vertically. A MACARONS module can be built to have any selected capacity.  Several modules can then also be co-located to scale up.

\begin{figure}[h!]
    \centering
    \includegraphics[width=0.4\linewidth]{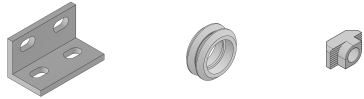}
    \caption{Example of common and readily available components: double L-bracket (left), V-wheel (middle) and drop-in tee-nuts (right).}
    \label{fig:common_parts}
\end{figure}

MACARONS is designed to be easily built out of cheap and easily available components, examples of these are shown in Figure \ref{fig:common_parts}. A stacked horizontal system is used due to its simpler geometry when compared to vertical geometries such as the green wall \cite{skyward}. The design considers both the vertical-farming nature of the system and the robotics, where the support structure both holds up the weight of the crop and provide rails to constrain the robot motion. The robot therefore only has to push the crop, which are on the wheeled carriages, without having to lift it. Consequently, the robot does not require powerful motors, nor does it require advanced navigation.

\subsubsection{Shelves}

\begin{figure}[h!]
    \centering
    \includegraphics[trim={20cm 0 3cm 0},clip, width=1.0\linewidth]{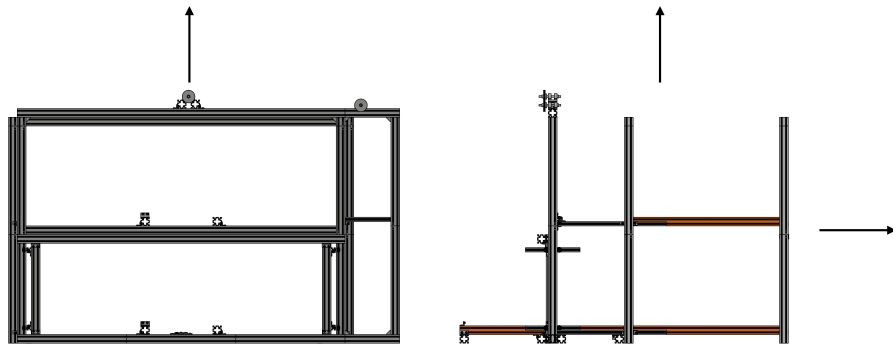}
    \caption{Front (left) and side (right) view of a minimal MACARONS with two vertical and one horizontal unit. The central rails (orange) support the weight of the crop, act as guide rails for the mover robot and is used by the robot to localise itself. The arrows indicate replication directions along the vertical and horizontal direction.}
    \label{fig:central_rail}
\end{figure}

A shelf holds $N_h$ grow units. It is  designed to be replicated and vertically stacked $N_{v}$ times as shown in Figure \ref{fig:central_rail}. For a $N_{h}\times N_{v}$ MACARONS, the shelves consist of $N_{h}\times N_{v}$ grow units. The V-slot aluminium profile lengths can be changed to accommodate available space and tray sizes, although this is not recommended. The aluminium profile lengths currently used are round number lengths (250mm, 500mm, 750mm and 1500mm) that can be dismantled and reused for other projects. To extend the standard lengths, double tee-nuts can be used. This allows for long continuous profiles, which are required for the V-wheels. Similarly, to make the profile ends fit nicely together, 40mm x 40mm x 40mm profiles can be added using double tee-nuts. Instead of having to custom cut profiles of specific lengths it is simpler from a logistics and re-usability point of view to have a large stock of round number lengths.

The central aluminium profile (Figure \ref{fig:central_rail}) is designed to be used as both the structural support and the guide rail for the mover robot and carriage. These are filled with V-slot fillers that are used for both supporting the mover motion and localisation using line-following infrared sensors. These are used for localisation, as opposed to tape which are typically used for line-following, as they do not come loose over time like tape, especially in potentially humid environments.

\subsubsection{Carriage}

\begin{figure}[h!]
    \centering
    \includegraphics[width=0.6\linewidth]{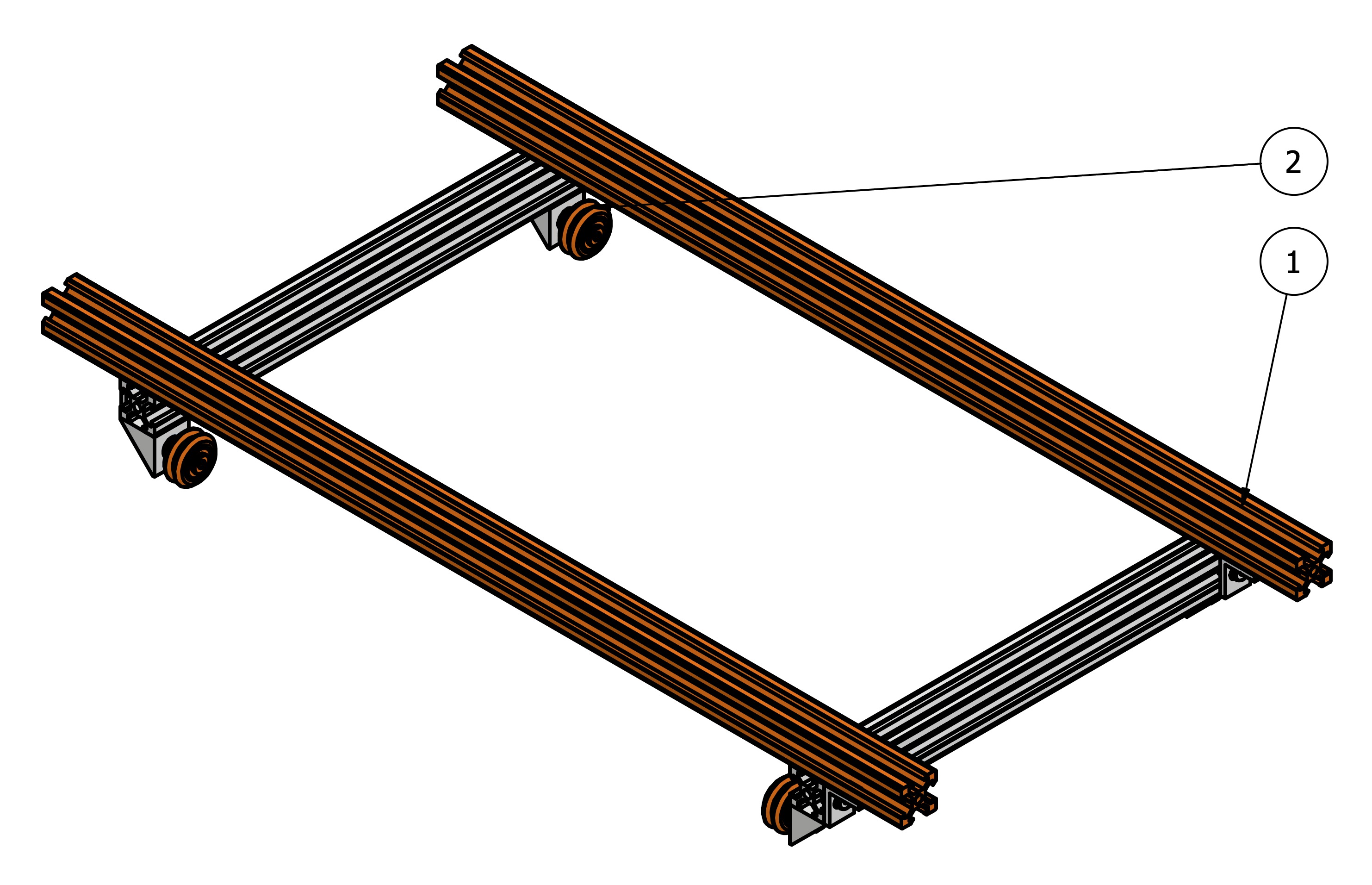}
    \caption{CAD image of a carriage. Annotated components are highlighted in orange. Aluminium extrusion (1) supports the payload. V-wheels (2) allows for smooth rolling of the payload.}
    \label{fig:carriage_picture}
\end{figure}

Every crop tray is permanently mounted to a passively wheeled carriage (Figure \ref{fig:carriage_picture}).  A $N_h \times N_v$ capacity module can contain up to $N_h \times N_v$ carriages. The carriage allows for the weight of the tray to be supported by the shelf, which lowers the specifications required by the mover (see below) for the translation of the trays. This increases the mover's cost-effectiveness, compactness and ease of material sourcing. The carriage is designed to be flexible and the brackets can be adjusted to fit and clamp on to trays of varying sizes. As opposed to using plastic rollers attached to the rails of the shelf, using a carriage is more reliable and easy to use.

\subsubsection{Mover}

\begin{figure}[h!]
    \centering
    \includegraphics[width=0.8\linewidth]{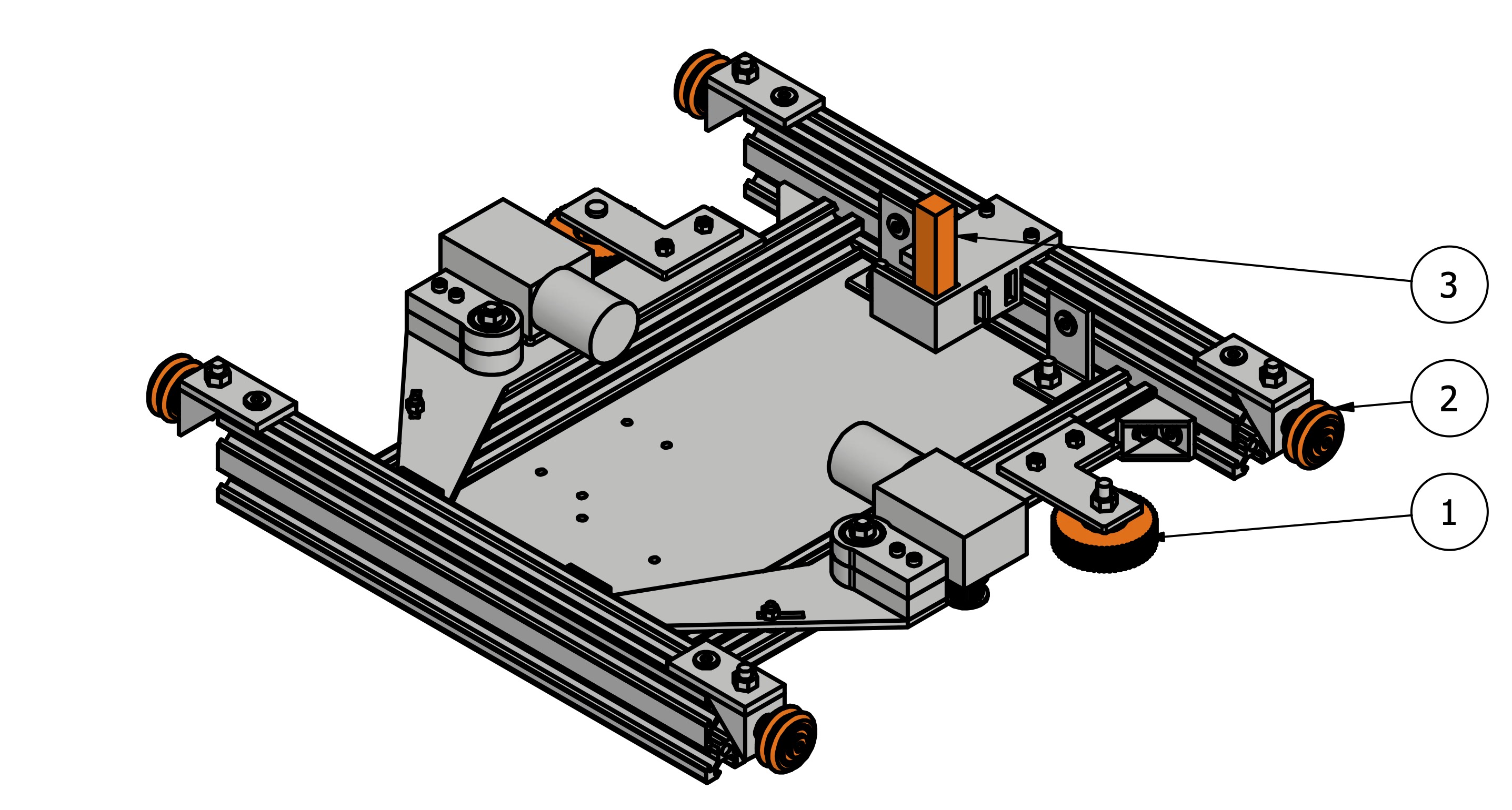}
    \caption{CAD image of the mover. Annotated components are highlighted in orange. The mover fits between guide rails. Spring-loaded driven wheels (1) push against the guide rails. V-wheels (2) slot into the guide rails holding the weight of the mover. Interaction arm (3) allows the mover to push/pull carriages along the guide rails.}
    \label{fig:mover_picture}
\end{figure}

The mover (Figure \ref{fig:mover_picture}) is an active robot that can attach to a carriage and transport it. There can be multiple movers in one MACARONS module, which will allow for simultaneous transportation of carriages. The mover is designed to be compact, fitting under the carriage. It uses 3D-printed wheels that engage with 2GT timing belts that are commonly used in 3D-printers. The 2GT timing belts are used to provide traction and to drive the wheels via 2GT pulleys and a 12V DC motor. This avoids the problem of finding appropriate wheels small enough to provide enough traction with the aluminium profiles. 

The two arms of the mover that hold the wheels are attached to each other via a tension spring. This allows the wheels to always be in good contact with the aluminium profile, ensuring that the mover is robust to potential misalignment in the aluminium profiles. The wheels engage with aluminium profile rails in a direction perpendicular to the weight of the mover. This also improves the robustness to derailing.

The mover is powered by one 12V DC power connector, allowing it to be easily powered by a 12V battery pack. This is useful if MACARONS is horizontally tessellated beyond one unit length. It is also useful for the operation of multiple movers within one MACARONS. The L298N is used as the motor driver as these are popular and readily available. The two 12V Geared DC motors used are commonly used in consumer products and come in a wide variety of specifications under one form factor.

\subsubsection{Elevator}

\begin{figure}[h!]
    \centering
    \includegraphics[width=0.8\linewidth]{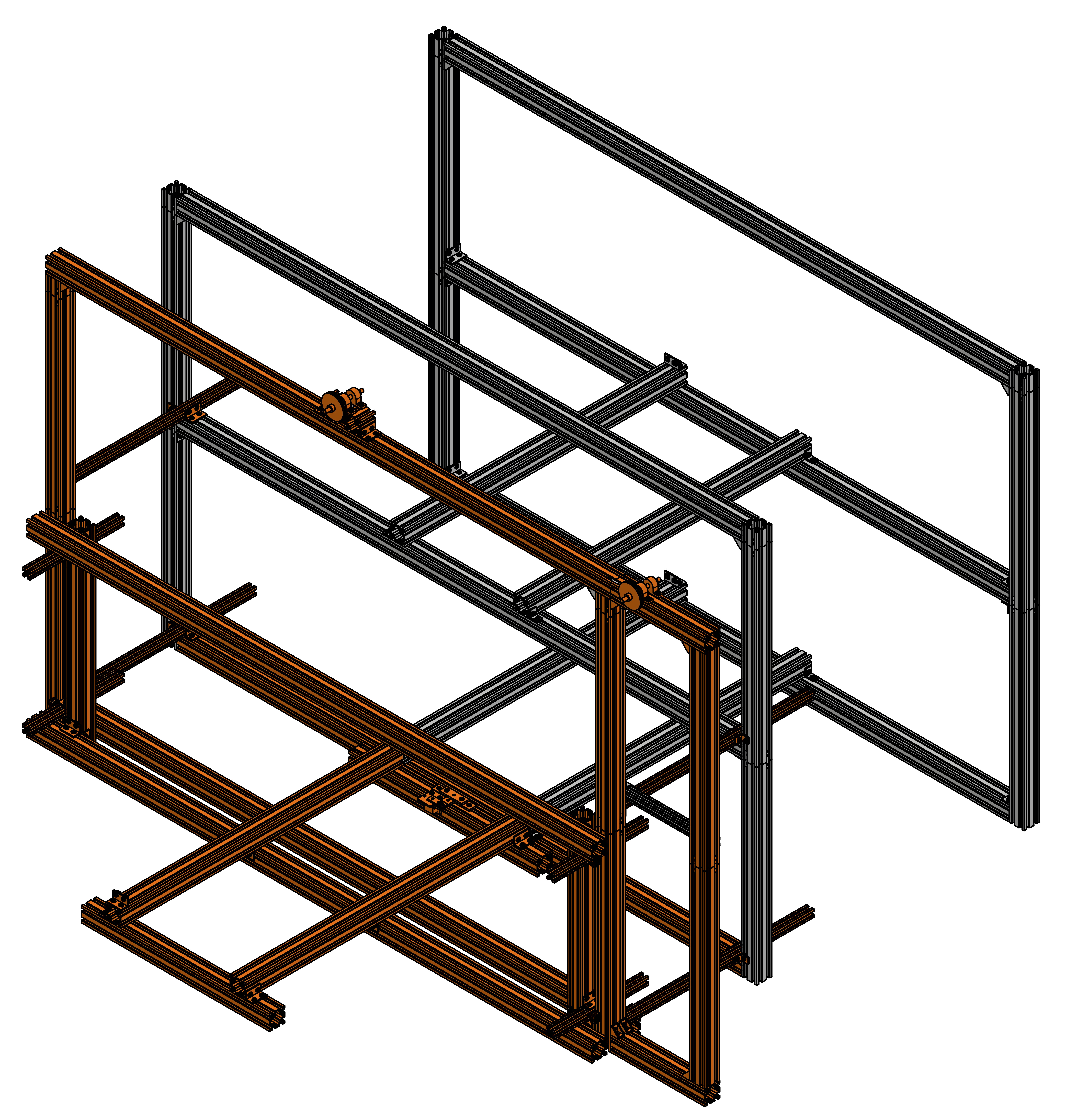}
    \caption{CAD image of the elevator (orange). The elevator allows for the mover and carriages to be moved between the different floors.}
    \label{fig:elevator_highlighted}
\end{figure}

The elevator (Figure \ref{fig:elevator_highlighted}) connects the shelves together and allows one carriage carried by a mover to be transported vertically between shelves. The elevator localisation is done using an infrared sensor which is cheap, readily available and does not require physical contact. It is simple to implement and easy to calibrate. The lifting of the elevator is implemented using an electric winch. These are readily available and comes in a variety of specifications and can be chosen for the specific needs of each MACARONS build. The electric winch, the motor driver used to drive it and the power supply must be compatible. As with the shelf, the elevator is designed to be replicated in the vertical direction. In the area above the winch, counterweights can also be added. This improves energy efficiency and allows for smaller winches to be used. The infrared sensor is placed on the elevator platform and has an adjustment that allows the sensor to be moved up and down, which is used to calibrate the sensor. By placing the sensor here we only require one sensor for the elevator and it only has to be calibrated once.

\subsection{Software}
Software is in Python and micro-Python and released under GPL3. Networking revolves around communication between the ESP32 (C3) microcontroller and a Raspberry Pi (4B) Server. The ESP32 (C3) is a cost-effective microcontroller with wifi, and is based on the open-source RISC-V architecture. 

In this framework, the devices function independently, copying configurations from the server to its local environment and using these configurations to execute tasks. The devices can push readings back to the server and multiple devices can be coordinated together to execute a task using scripts that are stored on the server. The website that is hosted by the web-server allows the user to add and reprogram these devices and scripts. It also allows the user to run the scripts, view the readings taken by the devices and update configurations. The device only has to be connected to the server once during the registration process. Once this is complete, all subsequent re-programming and file uploads can be done through the website over wifi.

Devices run the following sequence, looping forever.
\begin{enumerate}
    \item Wakes from deep sleep.
    \item Updates itself if there are staged updates on the webserver.
    \item Runs the main code.
    \item Updates itself if there are staged updates on the webserver.
    \item Enters deep sleep for a set period of time.
\end{enumerate}

This allows the device to enter energy saving modes periodically (deep sleep), which is useful for sensors that take readings periodically. It also allows for manual control of the device while it is awake.

The server holds the database for all the devices and scripts uploaded to it. It hosts the webserver that allows the user to access and manipulate the database through a website. Additionally, it coordinates the relevant devices when a script is ran and holds the information required to add devices such as the wifi credentials.

\section{Formal Specifications}
Specifications for the minimal ($2 \times 1$) MACARONS that has been  built are given below.

\subsection{General}

\begin{center}
\begin{tabular}{ |p{0.35\linewidth} | p{0.5\linewidth} |} 
 \hline
 Payload weight & 12.5 kg \\ 
 \hline
 Tray size & 1060mm $\times$ 630mm \\
 \hline
 Horizontal speed & 100mm/s  \\ 
 \hline
 Vertical speed & 41.7mm/s  \\ 
 \hline
 Approximate size & 1850mm $\times$ 1250mm $\times$ 1650mm\\
 \hline
 Elevator power supply & 12V DC, 30A, 360W\\
 \hline
 Elevator Input voltage & 90-132V or 180-264V AC (selected using a switch on the power supply.\\
 \hline
 Operating condition & Normal office room temperature and humidity in the UK.\\
 \hline
\end{tabular}
\end{center}

\subsection{Scaling}
Horizontal and vertical scaling can be easily achieved by replicating the design. The minimal MACARONS module built and tested includes two grow units stacked vertically and may potentially scale to higher vertical extent. To pack multiple MACARONS modules next to each other, the recommended distance between each MACARONS module is 1000mm to enable human access and maintenance.

\subsection{Website}
The number of devices that can be controlled remotely is two (mover and elevator) but may potentially support more. The website allows for the following:

\begin{itemize}
    \item Accessed through a web-browser.
    \item Manual control of the device configurations.
    \item Specify where to move grow tray or the mover using pre-uploaded scripts.
    \item Upload custom scripts.
    \item Upload custom mover software.
    \item Upload custom elevator software.
\end{itemize}

\subsection{Assembly}
Tools required are: hex keys, drill, soldering, screw driver, spanner, 3D-printer and laser-cutter. Time required for assembly is:

\begin{center}
\begin{tabular}{ |p{0.35\linewidth} | l |} 
 \hline
 Elevator and shelf & Up to 4 hours \\
 \hline
 Elevator peripherals & Up to 2 hours \\
 \hline
 Mover & Up to 1 hour\\
 \hline
 Component fabrication & Up to 6 hours\\
 \hline
\end{tabular}
\end{center}

\section{(3) Quality Control}\label{h.f8237gmzmwc6}
If a mechanical failure occurs, MACARONS could potentially damage itself, its crops and human operators. Electrical components also pose a risk to humans during building and operation. This section highlights features added to the design and to build instructions to mitigate these risks. Non-optional acceptance, integration and system tests are included in the build instructions. These respectively confirm that all components are functioning as required, that their installation is correct after key steps, and that new builds meet the specifications.

\subsection{Safety features}
{\em Universal L-brackets} (double) are placed mounted at the ends of the MACARONS guide rails to prevent the mover or carriages from falling off.

{\em Elevator lock}. During carriage transportation on the elevator, the mover interaction arm is locked on to prevent the carriage from falling off the elevator. 

All devices are independent from the server so will not crash if the {\em server crashes} mid-execution.

A {\em 20A fuse} is fitted to a fuse box between the elevator power supply and the motor driver inside the elevator control box to prevent damage to the system and humans if there is a fault such as a short circuit.  

An {\em enclosure} is used to house the power supply, motor driver, fuse box and ESP32 in the elevator control box to isolate the power electronics from the environment and humans.

\subsection{Testing}
Integration tests are included throughout the build to guarantee that  components are assembled correctly. The assembled carriage is used to test for smooth rolling along the rails. This ensure that all rail spacings are correct. A similar test is used to ensure proper alignment of the rails on the elevator platform and the shelf. To test that the elevator is aligned properly, a 3mm piece of plastic is placed between the elevator platform and the shelf edge. The V-wheels are adjusted accordingly. Mobility of the drag chain is tested by moving the elevator platform. Proper alignment of pulleys is ensured by testing the elevator cables are vertical. Structural integrity of the system is ensured by testing that all tee-nuts are locked in place. 

The elevator infrared sensor is tested by powering the elevator ESP32 and calibrating it. There are two green lights on the sensor, one indicating that the power is on and the other indicating that a reflection is detected. The detection light should turn on when the sensor is obstructed by an aluminium profile. The sensitivity of the sensor may need to be adjusted. The sensor's alignment is tested and calibrated by adjusting its location on the mount until the it is on the verge of triggering when the elevator platform and the shelf is aligned. 

Once a build is complete and the software uploaded onto the server/devices, a series of system tests are conducted to ensure that the formal specification is met.

\section{(4) Application}\label{h.f78bi3oom0mu}
\subsection{Vertical Farming}

For $N$ modules of $N_{h}$x$N_{v}$ MACARONS, the approximate total unloading time is given as follows:

\begin{equation}
    T_{mover} = N_{h}N_{v}\Bigl((N_{h}+1)t_{h}^{(m)} + (N_{v}+1)t_{v}^{(m)}\Bigr),
\end{equation}


where $t_{h}^{(m)}\approx$15s and $t_{v}^{(m)}\approx$12s is the time it takes the mover robot to move one unit horizontally and vertically, respectively. For a farm consisting of ten 10x10 MACARONS modules, this time is 8.3 hours.

Comparing this with one human unloading a vertical farm of comparable size, we assume that this operator takes no breaks and can lift heavy bulky trays onto the scissor lift with no error with an average time of $t_{m}\approx15s$. We assume the operator can unload all trays in a column onto the scissor lift before transporting it to the edge of the shelf (similar to the MACARONS transporting the trays to the elevator for unloading). For $N$ shelves the approximate unloading time is,
\begin{align*} 
T_{human} &= \Bigl(2N_{v}N_{h}t_{v}^{(s)} + N_{v}N_{h}t_{m} + N_{h}(N_{h} + 1)t_{h}^{(s)}\Bigr)N,          
\end{align*}
where $t_{h}^{(s)}\approx$10s, $t_{v}^{(s)}\approx$6s \cite{scissor_lift} is the time for the operator to move the scissor lift one unit horizontally and vertically, respectively. The first term captures vertical motion, the second is the unloading time and the third is  horizontal motion to deliver trays. For a farm of comparable size to ten 10x10 MACARONS modules, this time is 10.6 hours. 

It has been estimated that around 15\% of all labour is associated with moving trays around in vertical farms \cite{ifarm}; and that around 56\% of vertical farm operational costs are  labour \cite{pure_greens}. Estimated labour cost for vertical farms in the USA in 2020 was 222.18USD/m$^{2}$ \cite{pure_greens}. Using the total grow area of a vertical farm of comparable size in the above analysis, which is $217.5$m$^{2}$,  equating to around 48,325USD per year. Ten $10 \times 10$ MACARONS modules are therefore expected to save around 4110USD/year in labour.

\subsection{Reuse potential and adaptability}
MACARONS is designed to be easily modified. The sizes of the aluminium profile can be changed to match different space requirements, although the current dimensions are chosen to be round numbers. The carriages can be adjusted to clamp onto trays of varying sizes. As MACARONS is built out of aluminium profile, additional components such as sensors and actuators -- for example to monitor plant condition and apply water and fertilizers -- can be easily mounted anywhere on the platform.   It is also easy to scale MACARONS as it is designed to be replicated in the vertical and horizontal directions. Multiple alternatives also exist for the electronic components used. These can be switched out for open-sourced components when they become available. Adaptation and modification of the designs would require the same tools used in construction.

The scalability and capabilities of MACARONS could potentially be adapted for other contexts such as warehousing, lab-automation and as a low-cost alternative to conveyor-belt systems. Depending on the speed requirements different motors and motor drivers may have to be switched out. Multiple movers may also need to be included in the system. Although new software for the devices and scripts will need to be written, the software framework used is designed such that the software on each device is highly customisable and can be easily uploaded without change to the entire framework.

\section{(5) Build Details}\label{h.l8i9vokvs0bj}

\subsection{Availability of Materials and Methods}

The hardware designs are licensed under the CERN OSH-W and made from Available Components as under its definition. The majority of these are part of the OpenBuilds open-source hardware ecosystem which can be sourced from a number of suppliers worldwide such as OpenBuilds Part Store (and Ooznest in the UK). Other components are readily available  from multiple sellers. Custom parts are either laser-cut or 3D-printed. Note that all parts can be 3D-printed, but laser cutting was used as these required less time to manufacture. Total cost of the minimal demonstration build is 1535.50USD (2022).

\subsection{Ease of Build}
Mechanical assembly requires Allen keys, spanners, screw drivers, and soldering. For custom component fabrication, access to a 3D-printer, laser-cutter and drill is required.

\subsection{Operating Software and Peripherals}
Any device that can access the website and upload files can be used to fully control the system, remotely. The Raspberry Pi (4B) runs Raspbian Buster and the main language used is Python (3.8). The ESP32 (C3) micro-controllers are based on the open-source RISC-V architecture and run micro-Python (1.18). The 3D models and drawings can be edited using open-source FreeCAD software.

\subsection{Hardware Documentation and Files Location:}

\textbf{Archive for hardware documentation, build files and software}

\textbf{Name}: GitLab


\textbf{Project repository:} \url{https://gitlab.com/owopen-source/macarons}

\textbf{Licence:} CERN-OHL-W hardware design and build instructions; GPL3 software source code; CC-BY4 additional documentation. 

\textbf{Date published:} 2022-10-10

\section{(6) Discussion}\label{h.90jl7wm65t65}

A minimal MACARONS (two units high and one unit deep) was constructed and tested. It is designed to move horizontally along the guide rails at 100mm/s and up the elevators at 41.7mm/s. The system can successfully carry trays of plants up to 1060$\times$630mm$^{2}$ and 12.5kg. A simple analyses demonstrated that MACARONS can operate at a rate sufficient to automate tray loading/unloading and hence reduce labour costs. 

Future works to extend MACARONS include the incorporation of lights and irrigation to provide a fully functional automated farming system. Currently MACARONS can only load and unload sequentially. A variation that allows for access to any crop tray at any time will be developed. Once complete MACARONS will be tested by growing microgreens. 

Many components in MACARONS are not open-sourced. A future direction would be to replace these components with open-sourced counterparts when these arise or to design these components ourselves.  Open-source wheels from the OpenWheel project \cite{openwheel} could be included when that project completes.

More automation could also be added, such as cameras and nutrient sensors for monitoring and an interface that connects to the elevator allowing for the crop to be unloaded to other stations such automated harvesting and seeding stations.

Most commercial greenhouses \cite{cucumber_humidity, leafy_greens_humidity} operate within similar humidity ranges to the office levels currently tested \cite{office_humidity}. Some specialist crops however may grow more optimally in higher humidity environments.  Versions of MACARONS for use in such environments could be forked in the future, specifically by replacing the motors and motor drivers with suitable IP rated, but higher-cost, alternatives.  All other potentially sensitive components are already enclosed in an IP56 enclosure which is sufficient for these environments. 

Further work must be conducted to optimise vertical farms to improve their economic viability.  MACARONS now provides a platform to accelerate and accumulate its progress, and facilitate collaboration in vertical farming.

\section*{Paper author contributions}\label{h.fy8hbipy6kwe}
Vijja Wichitwechkarn designed and built the system under the supervision of Charles Fox. Both authors contributed to the paper text.

\section*{Acknowledgements}\label{h.gu3yyarx72d6}
Thanks to Chris Waltham for guidance with mechanical assembly and electronics, and Rob Lloyd for advice and help with component fabrication.

\section*{Funding statement}\label{h.4u1a7tugh2om}
This work was supported by the Engineering and Physical Sciences Research Council [EP/S023917/1]

\section*{Competing interests}\label{h.q1j1rznb43fl}

    The authors declare that they have no competing interests.

\bibliographystyle{unsrt}
\bibliography{macarons}

\section*{Copyright notice}\label{h.jm5gcqv4g8x0}
\copyright~2022 The Authors. This is an open-access article distributed under the terms of the Creative Commons Attribution 4.0 International License (CC-BY 4.0), which permits unrestricted use, distribution, and reproduction in any medium, provided the original author and source are credited. See http://creativecommons.org/ licenses/by/4.0/.

\end{document}